\documentclass[a4paper,conference]{IEEEtran}
\usepackage{cite}

\ifCLASSINFOpdf
  \usepackage[pdftex]{graphicx}
\else
  \usepackage[dvips]{graphicx}
\fi
\usepackage{amsmath}
\usepackage{algorithmic}
\ifCLASSOPTIONcompsoc
  \usepackage[caption=false,font=normalsize,labelfont=sf,textfont=sf]{subfig}
\else
  \usepackage[caption=false,font=footnotesize]{subfig}
\fi
\usepackage{amssymb}  
\usepackage{xcolor}
\usepackage{multirow}
\usepackage{enumitem}

\definecolor{myred}{HTML}{aa0000}
\definecolor{myorange}{HTML}{ff7700}
\definecolor{mygreen}{HTML}{00aa00}
\definecolor{myblue}{HTML}{0000cc}

\begin{document}
%
\title{\textbf{S}ubjective \textbf{A}ssessments of \textbf{L}egibility in \textbf{A}ncient \textbf{M}anuscript \textbf{I}mages - The SALAMI Dataset}

\author{\IEEEauthorblockN{Simon Brenner}
\IEEEauthorblockA{Computer Vision Lab,\\Institute of Visual Computing and\\Human-Centered technology\\TU Wien\\Vienna, Austria\\Email: sbrenner@cvl.tuwien.ac.at}
\and
\IEEEauthorblockN{Robert Sablatnig}
\IEEEauthorblockA{Computer Vision Lab,\\Institute of Visual Computing and\\Human-Centered technology\\TU Wien\\Vienna, Austria\\Email: sab@cvl.tuwien.ac.at}}


%


\maketitle

\begin{abstract}
The research field concerned with the digital restoration of degraded written heritage lacks a quantitative metric for evaluating its results, which prevents the comparison of relevant methods on large datasets. 
Thus, we introduce a novel dataset of Subjective Assessments of Legibility in Ancient Manuscript Images (SALAMI) to serve as a ground truth for the development of quantitative evaluation metrics in the field of digital text restoration.
This dataset consists of 250 images of 50 manuscript regions with corresponding spatial maps of mean legibility and uncertainty, which are based on a study conducted with 20 experts of philology and paleography. As this study is the first of its kind, the validity and reliability of its design and the results obtained are motivated statistically: we report a high intra- and inter-rater agreement and show that the bulk of variation in the scores is introduced by the images regions observed and not by controlled or uncontrolled properties of participants and test environments, thus concluding that the legibility scores measured are valid attributes of the underlying images.
\end{abstract}


%
\IEEEpeerreviewmaketitle

\section{Introduction}
\label{sec:introduction}

Written heritage is a valuable resource for historians and linguists. However, the physical medium preserved may be in a condition that prohibits the direct accessing of the text. Addressing this problem, a research field dedicated to the digital restoration of such degraded sources based on specialized imaging techniques, such as multi- and hyperspectral imaging as well as x-ray fluorescence mapping, has ensued~\cite{Arsene2018, Easton2011, Giacometti2017, Glaser2016, Hollaus2015, Mindermann2018, Pouyet2017, Salerno2007}. In general, corresponding approaches aim at producing output images in which a text of interest is maximally legible for a human observer.

We note that an inherent problem of this research field is the absence of a suitable quantitative metric for this property of human legibility. Consequently, the evaluation of proposed approaches is commonly based on expert ratings, the demonstration on selected examples or case studies~\cite{Easton2011, Hollaus2015, Mindermann2018, Pouyet2017, Salerno2007}. This practice is unfavorable for the research field: it does not allow for an automated evaluation on large public datasets, such that an objective comparison of different approaches is impeded.

Attempts to quantitatively assess the success of legibility enhancements have been made before: Arsene et al. quantify the success of dimensionality reduction methods via cluster separability metrics, however acknowledging that the resulting scores do not correlate well with human assessments~\cite{Arsene2018}. Shaus et al. introduce the metric of "potential contrast" between user-defined foreground and background areas~\cite{Shaus2017}.
Giacometti et al. created a multispectral image dataset of manuscript patches before and after artificial degradation~\cite{Giacometti2017}, which allows the quantitative assessment of digital restoration approaches by comparison with the non-degraded originals. 
The performance of Handwritten Character Recognition (HCR) systems on the enhanced images is used as a quantitative metric as well~\cite{Hollaus2014, Likforman-Sulem2011}. This approach addresses the property of readability more directly than the approaches mentioned before and is a reasonable choice when the purpose of text restoration lies in the subsequent processing with a HCR system, rather than in the preparation for a human observer.
For restoration approaches producing binary images, evaluation is more straight-forward: for this purpose, multispectral image datasets with ground truth annotations have been published~\cite{Hedjam2015, Hollaus2019}.
While the above-named approaches to quantitative evaluation are promising for their respective use cases, their correlation to legibility by human observers is yet to be shown.

For the development of general objective Image Quality Assessment (IQA) methods, the use of public databases containing subjectively rated images is a well-established practice~\cite{CID2013,Perez-Ortiz2019}. A variety of such datasets have been published; they primarily aim at measuring the perceptual impacts of technical parameters such as image compression artefacts, transmission errors, or sensor quality~\cite{LIVE,TID2008,TID2013,CID2013,KADID10k}. However, no such dataset exists for the assessment of text legibility in images.

This paper introduces a new dataset of manuscript images subjectively rated for legibility, designed for the development and validation of objective legibility assessment methods. The main contributions of this work are:

\begin{itemize}
	\item a methodology to conduct studies of subjective legibility assessment in manuscript images.
	\item publication of a novel dataset that serves as a reference for the development of objective legibility assessment methods.
\end{itemize}

The remainder of the paper is structured as follows:
Section~\ref{sec:design} describes the methodology of the subjective IQA study carried out, while Section~\ref{sec:conduction} details its technical and practical implementation. In Section~\ref{sec:evaluation} we analyze the results obtained, motivating the validity of our study design and the properties of the published dataset. Section~\ref{sec:dataset} describes the published dataset in detail, and Section~\ref{sec:conclusion} concludes the paper with final observations and potentials for improvement.


\section{Study Design}
\label{sec:design}
In the following, the design of the subjective IQA study carried out to establish the SALAMI dataset is described and motivated. The documents ITU-T P.910~\cite{ITU2008} and ITU-R BT.500-13~\cite{ITU2012} by the International Telecommunication Union provide guidelines for the implementation of subjective IQA studies that are commonly followed in the creation of respective datasets~\cite{Ribeiro2011, Ghadiyaram2016, CID2013, DeSimone2009, KADID10k}. In the design of this study, we implement these guidelines wherever applicable.

\subsection{Test images}
\label{sec:testimages}

Our test image set is based on 50 manuscript regions of 60x60mm, each of which is represented by 5 images. The regions are sampled from 48 different manuscripts and 8 language families. Slavonic (19), Latin (13) and Greek (12) texts make up the majority of the samples; additionally, two Ottoman texts and one each in Armenian, Georgian, German and Gothic are contained in the dataset. Depending on line height and layout, the regions contain 1-17 lines of text. In the following, the selection of manuscript regions and the creation of the final image set is described in detail, and choices made in the process are motivated.

The manuscript regions represented in the SALAMI dataset are drawn from a set of approximately 4600 pages of 67 historical manuscripts, of which the Computer Vision Lab (TU Wien) has acquired multispectral images in the course of consecutive research projects between 2007 and 2019\footnote{Projects financed by the Austrian Science Fund (FWF) with grant numbers P19608-G12 (2007-2010), P23133 (2011-2014) and P29892 (2017-2019), as well as a project financed by the Austrian Federal Ministry of Science, Research and Economy (2014-2016)}. The multispectral images were acquired using different imaging devices and protocols~\cite{Diem2008, Hollaus2012, Hollaus2019}, with 6-12 wavebands per image.

Rather than presenting whole manuscript pages to the participants, we reduce the image contents to square-shaped regions of 60mm side-length. This enables the presentation of a fixed region in sufficient magnification, avoiding the need for zooming and panning. The assessment task is thus simplified and accelerated, and the risk of overlooking small pieces of text is minimized.
To limit the scope of this study, manuscript regions containing multiple layers of text (such as palimpsests or interlinear glosses) are excluded and must be considered in future work.

In order to select 50 suitable regions from the entirety of available pages, a semi-automatic scheme of random selection and human redaction was employed. It aims at minimizing human bias while ensuring that selected regions contain only a single layer of text. At the same time, the variety of manuscripts from which the samples are drawn is maximized.
In a first step, 50 suitable pages are selected: from each of the available manuscripts, a randomly drawn page is presented to an operator, who accepts the image if it contains at least a 60x60mm area with exactly one layer of text present (where the text is not required to cover the whole area), and rejects it otherwise. In the next iterations, only manuscripts from which no pages were accepted previously are considered. The process is repeated until (a) a page  of each manuscript has been accepted, (b) all pages from the remaining manuscripts have been rejected or (c) the target amount of 50 pages has been reached. In cases (a) and (b) the iteration is restarted, with unvetted pages from all manuscripts available again. 
In the second step, a similar strategy is used to select the final regions from the drawn images: for each image, randomly selected 60x60mm regions are presented to the operator sequentially, until one is accepted.

%

The source multispectral images are cropped to the selected square regions and re-sampled to a standard resolution of 12 px/mm (or 304.8 ppi), resulting in an image size of 720x720 pixels. This standardization is done to eliminate image size and resolution as a source of variance.

\begin{figure} 
	\centering
	\subfloat[]{%
		\includegraphics[width=1.0\linewidth]{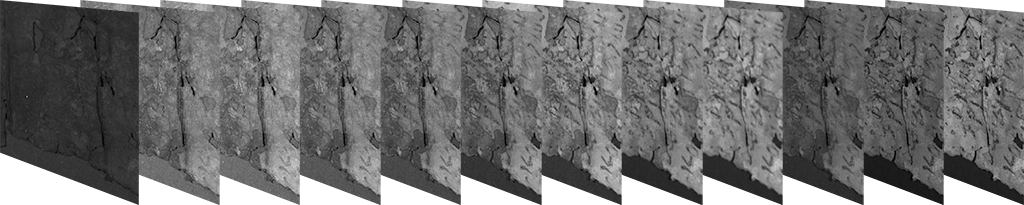}
		\label{fig:example:a}
	}\\
	\subfloat[]{%
		\includegraphics[width=1.0\linewidth]{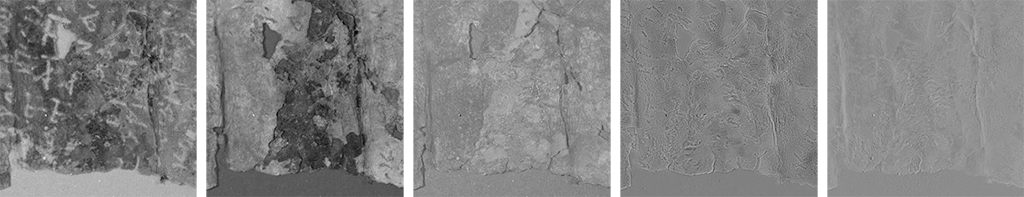}
		\label{fig:example:b}
	}
	\caption{Test image generation: (a) multispectral source layers of a given manuscript region. (b) the principal components corresponding to the 5 largest eigenvalues constitute the published test images.}
	\label{fig:example} 
\end{figure}

For each region, five processed variants are generated to serve as the actual test images. For the sake of simplicity and repeatability, those variants are produced by a principal component analysis on the multispectral layers\footnote{PCA is frequently used as a standard procedure for dimensionality reduction and source separation in multispectral manuscript images~\cite{Arsene2018, Giacometti2017, Mindermann2018}}. Figure~\ref{fig:example} shows an example.
The inclusion of multiple versions of the same manuscript region enables a versatile use of the SALAMI dataset: additionally to absolute rating applications, it can be used for the development of systems in which relative comparisons between multiple images of the same content are paramount.
With 5 variants for each of the 50 manuscript regions the SALAMI dataset contains a total of 250 test images. According to preliminary tests, this number of images can be assessed in approximately one hour by a single participant.

\subsection{Test method}

We subclassify the test methods commonly used for performing IQA studies~\cite{ITU2008, ITU2012, Perez-Ortiz2019, LIVE, Mantiuk2012} into three basic categories:

\begin{enumerate}
	\item \textbf{Absolute Rating}: the participant assigns an absolute quality score to each test image, without a reference image.
	\item \textbf{Degradation Rating}: the test variants are presented together with an optimal reference image of the same content. The participant rates the deviation of the test image from the reference.
	\item \textbf{Pair Comparison:} Relative scores are obtained by comparing pairs of test variants.
\end{enumerate}

Approaches that make use of comparisons to a reference image are advantageous for experiments, in which degradations of an optimal original are evaluated (such as JPEG compression~\cite{LIVE}). In our case such an optimal reference does not exist, such that this class of methods is not applicable.

Pair Comparisons between variants of the same content are shown to provide higher discriminatory power and lower variance than Absolute Ratings, especially when the perceptual differences between those variants are small~\cite{TID2008, ITU2008, Mantiuk2012}. However, the downsides of this approach are the high number of necessary comparisons ($({n^2-n})/{2}$ for $n$ variants) and the lack of a common absolute scale among different contents. The latter problem is solvable when performing cross-content comparisons~\cite{Perez-Ortiz2019}, however, it is not clear if such direct comparisons between different manuscript regions (which vary in preservational condition, size and amount of text and alphabets) are meaningful.

For this study, we opt for Absolute Ratings as a base design. Additionally to the above named reasons, this approach readily allows the following extension for a more detailed specification of legibility: instead of asking the participants to assign a single score to the whole test image, they are required to mark all visible text with rectangular bounding boxes, which are then rated individually (see Figure~\ref{fig:gui}). With this approach we obtain a spatial distribution of legibility instead of a single score per image.

\subsection{Rating scale}

Following ITU recommendations~\cite{ITU2012}, a five point rating scale is used. As we only want to assess text legibility and not any other quality of the image, we refrain from using the standard category labels given by ITU ('Excellent', 'Good', 'Fair',...); they could lead to misinterpretations of the task. Instead, we explicitly break down the property of legibility to the percentage of text within a given area that a participant deems clear enough to read. Dividing the available range into 5 equal intervals leads to the labels and corresponding numerical scores shown at the top of Table~\ref{tab:ratingoptions}. With the phrasing of legibility in terms of percentages we obtain scores on a true interval scale, which can not be assumed for the usual qualitative category descriptions~\cite{Ye2013}.

\begin{table}[!t]
	\renewcommand{\arraystretch}{1.1}
	\caption{Top: available rating options. Bottom: options for participant self-evaluation.}
	\label{tab:ratingoptions}
	\centering
	\begin{tabular}{|c|l|l|}
		\hline
		\textbf{property} & \textbf{verbal description} & \textbf{numerical value}\\
		\hline\hline
		\multirow{5}{*}{\rotatebox[origin=c]{90}{\shortstack{legibility \\ score}}} 
		& 80-100\% readable & 5\\
		& 60-80\% readable & 4\\
		& 40-60\% readable & 3\\
		& 20-40\% readable & 2\\
		& 0-20\% readable & 1\\
		& [non-selected areas] & 0\\
		\hline \hline
		\multirow{4}{*}{\rotatebox[origin=c]{90}{\shortstack{language \\ expertise}}} 
		& expert (primary research field)& 3\\
		& advanced (can read and understand) & 2\\
		& basics (knows the script) & 1\\
		& unacquainted & 0 \\
		\hline
		\multirow{4}{*}{\rotatebox[origin=c]{90}{\shortstack{SMI \\ exposure}}} 
		& multiple times a week & 3\\
		& multiple times a month & 2\\
		& occasionally & 1\\
		& never & 0 \\
		\hline
		\multirow{4}{*}{\rotatebox[origin=c]{90}{\shortstack{academic \\ level}}} & professor & 4 \\
		& post-doc & 3\\
		& pre-doc & 2\\
		& student & 1\\
		& none & 0\\
		\hline
	\end{tabular}
\end{table}

\subsection{Test environment}
Traditionally, subjective IQA experiments are carried out under controlled laboratory conditions~\cite{ITU2008, ITU2012}. However, a study by Ribeiro et al.~\cite{Ribeiro2011} shows that subjective ratings on the LIVE~\cite{LIVE} dataset obtained in laboratory conditions can be accurately reproduced in crowd sourcing experiments conducted with Amazon Mechanical Turk. Ghadiyaram and Bovik~\cite{Ghadiyaram2016} create a IQA dataset of 1162 mobile camera images rated by over 8100 participants online and report excellent internal consistency.

Considering these results and our special requirements for the participants (see Section~\ref{sec:participants}) that lead to a relative shortage of suitable volunteers, we decided to loosen the laboratory constraints and allow both participation in controlled conditions and on-line participation. A statistical comparison between the results of laboratory and online participation is given in Section~\ref{sec:evaluation}.

Participants were allowed to ask questions during the instruction and tutorial phases. Online participants could make use of this option via email or phone.

\subsection{Order of presentation}
\label{sec:orderofpresentation}

The test images are divided into five batches, where each batch contains one variant of each manuscript region. The assignment of the individual variants to the batches is done randomly, but equally for all participants. Within those batches, the images are randomly shuffled for each participant individually. In order to measure intra-rater variability, one randomly chosen image per batch is duplicated.

\subsection{Participants}
\label{sec:participants}

ITU-T P.910 recommends a minimum of 15 participants for any IQA study, while stating that four participants are the absolute minimum that allows a statistical assessment and there is no use in having more than 40 participants~\cite{ITU2008}.

For the study we recruit 20 participants among researchers in the fields of philology and paleography, that have experience in reading original manuscripts. We look for a mixture of university students, pre-doctoral researchers, post-doctoral researchers and professors.

\section{Experiment Conduction}
\label{sec:conduction}
In the following sections details about the implementation and conduction of the study are given.

\subsection{User interface}
For an efficient and consistent conduction of the experiment we provide a web-based user interface for the assessment task, which is equally used by participants in laboratory conditions and online participants.

\begin{figure} 
	\centering
	\includegraphics[width=1.0\linewidth]{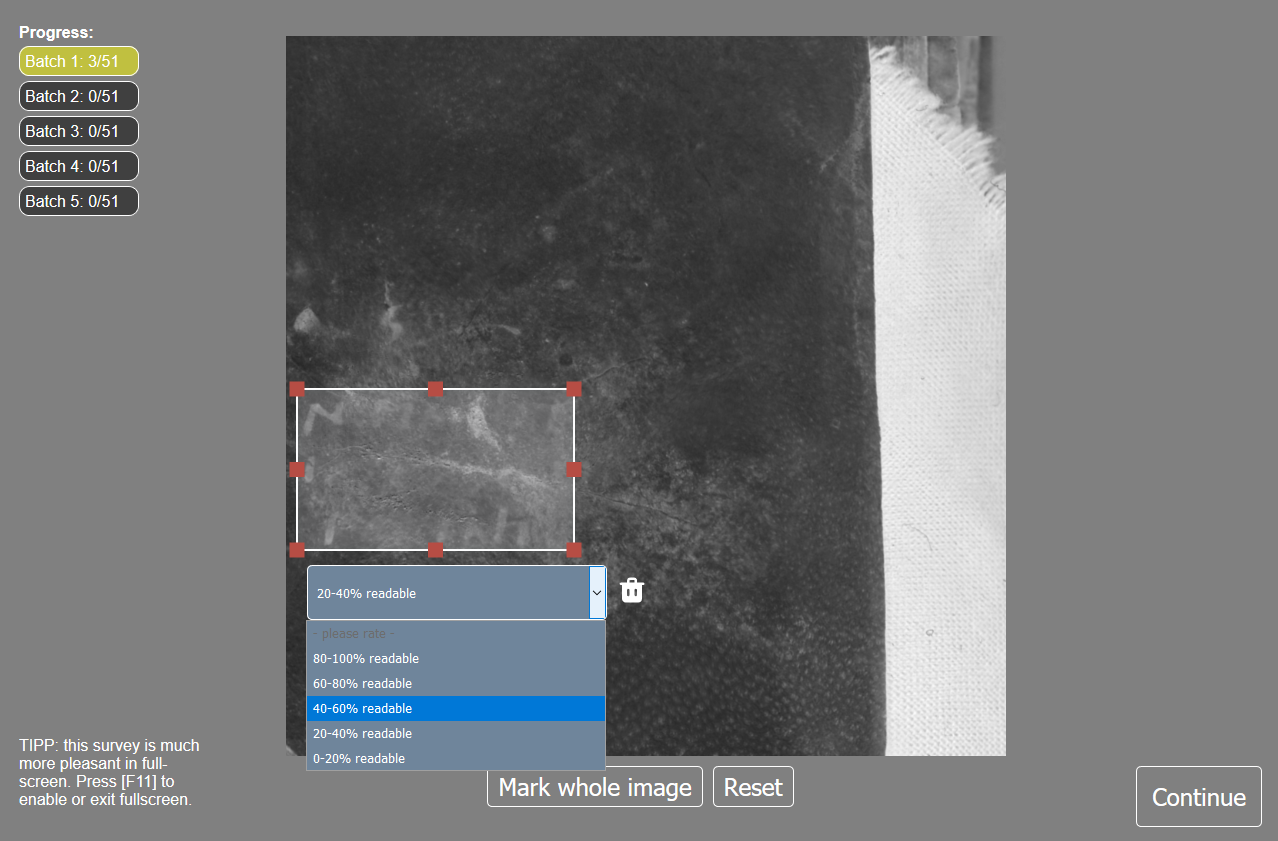}
	\caption{A screenshot of the user interface for legibility assessment. The image is displayed on a neutral gray background. The participant is required to mark blocks of visible text with bounding boxes and give an estimation on how much of this text can be read.}
	\label{fig:gui}
\end{figure}

During the primary test one image at a time is displayed on a medium gray background (50\% brightness, following ITU-T P.910 recommendations~\cite{ITU2008}). The participant is required to mark text areas with approximate bounding boxes and individually rate them. For this rating, the estimated amount of legible text within the marked region is chosen from a list, according to Table~\ref{tab:ratingoptions}. Figure~\ref{fig:gui} shows a screenshot of the primary test interface.

Prior to performing the primary test, the participants must complete three preparatory stages:
\begin{enumerate}
	\item \textbf{Self-assessment.} Participants are required to answer questions about their professional background: academic level, expertise in each of the language families present in our dataset and frequency of exposure to scientific manuscript images (from here on referred to as \textit{SMI exposure}) are queried. The available options for those questions are listed at the bottom part of Table~\ref{tab:ratingoptions}.
	\item \textbf{Instructions.} Participants are presented a sequence of pages, in which their task and the functionality of the user interface are explained, each supported by a demonstrative animation.
	\item \textbf{Tutorial.} 5 images are assessed within the primary test interface without the answers being recorded. These images were manually selected to cover a variety of cases: 
	\begin{itemize}
		\item a perfectly legible text covering the whole image
		\item two examples of faint text covering limited areas of the image
		\item a text of complex spatial distribution, encouraging the use of multiple bounding boxes
		\item an image without a visible text
	\end{itemize}
	These tutorial images were selected from batch 5, in order to maximize the distance between their occurrence in the tutorial and the primary test.
\end{enumerate}

\subsection{Laboratory Environment}

All laboratory test sessions were conducted on the same workplace using a Samsung SyncMaster 2493HM LCD monitor with a screen diagonal of 24 inches at a resolution of 1920x1200 pixels, and a peak luminance of 400cd/m$^{2}$. Viewing distance was not restricted, as this would not reflect a real situation of manuscript studying.

All laboratory participants rated the full set of test images (including duplicates) in a single session. They were allowed to take breaks at any time; however, none of the participants made use of this option.

\section{Evaluation}
\label{sec:evaluation}

In total, 4718 assessments were obtained from 20 participants (excluding duplicates for intra-rater variability). Not every participant rated all of the 250 test images, as some of the online-participants terminated the test earlier. The median time required to assess a single image was 12.4 seconds (with quartiles $Q_1=7.5s$ and $Q_3=22.4s$).

As the participants were free to select arbitrary image areas for rating, those areas can not be directly related between participants. Instead, each assessment (a given image rated by a given participant) is interpreted as a score map which is zero in non-selected areas and a value in the range from 1 to 5 in selected areas, according to their rating. Intersection areas of overlapping bounding boxes receive their maximum score. The choice of the maximum (over a median or rounded mean) was motivated by the observation that laboratory participants deliberately placing one bounding box on top of another always intended to label small areas with higher legibility than their surroundings, and never the other way around.

For the following analysis we partition the images in observational \textit{units} of 2x2mm (corresponding to 24x24 pixels). An elementary \textit{observation} is defined as the rounded mean of scores assigned to the pixels of a given unit by a single participant; accordingly, an image assessed by one participant results in 900 observations.

\begin{figure} 
	\centering
	\subfloat[]{%
		\includegraphics[width=0.49\linewidth]{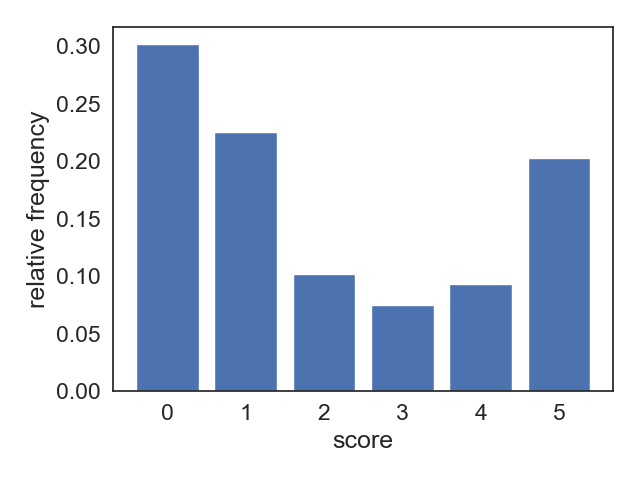}
		\label{fig:rawdata:hist}
	}
	\subfloat[]{%
		\includegraphics[width=0.49\linewidth]{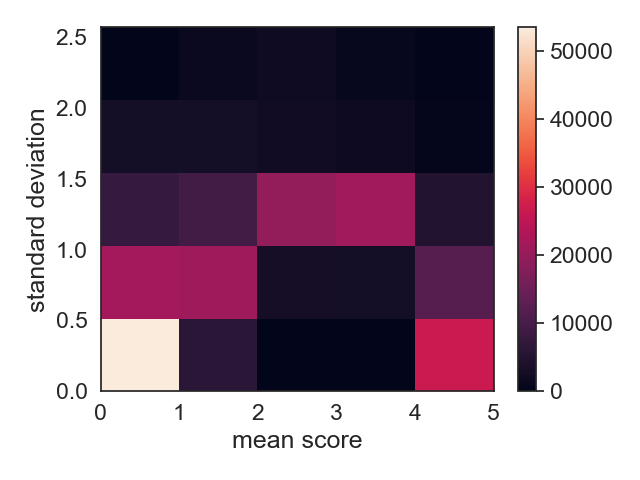}
		\label{fig:rawdata:hist2d}
	}
	\caption{(a) relative score frequencies over all observations; (b) 2D histogram of mean scores and standard deviation of all units.}
	\label{fig:rawdata} 
\end{figure}

A first look at the relative frequencies of scores over all observations (see Figure~\ref{fig:rawdata:hist}) leads to the insight that the bulk of observations report background, i.e. units where no text is visible. Furthermore, the 2D histogram of unit-wise mean scores and standard deviations shows a concentration of units with low scores and low standard deviations. This suggests that participants largely agree on the distinction between foreground and background. In order to prevent this mass of trivial background observations from biasing our analysis of intra- and inter-participant agreement, we exclude units that are labeled as background by more than half of the participants from all statistical considerations of this section.

\subsection{Participant characteristics and agreement}

Participant screening was performed following the algorithm described in ITU-R BT.500~\cite{ITU2012}. None of the participants were rejected. Table~\ref{tab:participants} shows the self-assessed properties of all 20 participants, where the numbers refer to the verbal descriptions given in Table~\ref{tab:ratingoptions}.

\begin{table}[!t]
	\renewcommand{\arraystretch}{1.1}
	\setlength{\tabcolsep}{1.2pt}
	\caption{Participants overview. Verbal descriptions of the numerical values are given in Table~\ref{tab:ratingoptions}.}
	\label{tab:participants}
	\centering
	 \fontsize{8}{8}\selectfont
\begin{tabular}{|r|rrrrrrrrrrrrrrrrrrrr|}
	\hline
	ID & \ 0& \ 1& \ 2& \ 3& \ 4& \ 5& \ 6& \ 7& \ 8& \ 9& 10& 11& 12& 13& 14& 15& 16& 17& 18& 19\\
	\hline
academic level  & \textcolor{mygreen}{2}  & \textcolor{mygreen}{2}  & \textcolor{mygreen}{2}  & \textcolor{mygreen}{2}  & \textcolor{myred}{4}    & \textcolor{mygreen}{2}  & \textcolor{myred}{4}    & \textcolor{myred}{4}    & \textcolor{mygreen}{2}  & \textcolor{myblue}{1}   & \textcolor{myblue}{1}   & \textcolor{mygreen}{2}  & \textcolor{myred}{4}    & \textcolor{myred}{4}    & \textcolor{myorange}{3} & \textcolor{myred}{4}    & \textcolor{myorange}{3} & \textcolor{myorange}{3} & \textcolor{myred}{4}    & \textcolor{myorange}{3}\\
SMI exposure & \textcolor{myorange}{3} & \textcolor{myorange}{3} & \textcolor{myorange}{3} & \textcolor{myblue}{1}   & \textcolor{myorange}{3} & \textcolor{myorange}{3} & \textcolor{mygreen}{2}  & \textcolor{myorange}{3} & \textcolor{myorange}{3} & \textcolor{myorange}{3} & \textcolor{myorange}{3} & \textcolor{myorange}{3} & \textcolor{mygreen}{2}  & \textcolor{myorange}{3} & \textcolor{myblue}{1}   & \textcolor{myorange}{3} & \textcolor{myorange}{3} & \textcolor{mygreen}{2}  & \textcolor{mygreen}{2}  & \textcolor{myorange}{3}\\ \hline
Armenian   & 0                       & 0                       & 0                       & 0                       & 0                       & 0                       & 0                       & 0                       & 0                       & 0                       & 0                       & 0                       & 0                       & 0                       & 0                       & 0                       & 0                       & 0                       & 0                       & 0\\
Georgian   & 0                       & 0                       & 0                       & 0                       & 0                       & 0                       & 0                       & 0                       & 0                       & 0                       & 0                       & 0                       & 0                       & 0                       & 0                       & 0                       & 0                       & 0                       & 0                       & 0\\
German     & \textcolor{myorange}{3} & \textcolor{myorange}{3} & \textcolor{mygreen}{2}  & \textcolor{mygreen}{2}  & \textcolor{mygreen}{2}  & \textcolor{myblue}{1}   & \textcolor{mygreen}{2}  & \textcolor{mygreen}{2}  & \textcolor{myorange}{3} & \textcolor{myorange}{3} & \textcolor{myorange}{3} & \textcolor{myorange}{3} & \textcolor{mygreen}{2}  & \textcolor{mygreen}{2}  & \textcolor{mygreen}{2}  & 0                       & \textcolor{myblue}{1}   & \textcolor{myorange}{3} & \textcolor{mygreen}{2}  & \textcolor{myorange}{3}\\
Gothic     & 0                       & \textcolor{myblue}{1}   & 0                       & 0                       & 0                       & 0                       & 0                       & 0                       & 0                       & 0                       & \textcolor{myblue}{1}   & 0                       & \textcolor{myblue}{1}   & \textcolor{myorange}{3} & 0                       & 0                       & 0                       & 0                       & 0                       & 0\\
Greek      & \textcolor{myblue}{1}   & 0                       & \textcolor{mygreen}{2}  & 0                       & \textcolor{myblue}{1}   & \textcolor{myorange}{3} & \textcolor{myblue}{1}   & \textcolor{mygreen}{2}  & 0                       & 0                       & \textcolor{myblue}{1}   & 0                       & \textcolor{mygreen}{2}  & \textcolor{mygreen}{2}  & \textcolor{myblue}{1}   & \textcolor{myorange}{3} & \textcolor{myorange}{3} & \textcolor{myblue}{1}   & \textcolor{myorange}{3} & \textcolor{mygreen}{2}\\
Latin      & \textcolor{mygreen}{2}  & \textcolor{myblue}{1}   & \textcolor{mygreen}{2}  & 0                       & \textcolor{mygreen}{2}  & \textcolor{mygreen}{2}  & \textcolor{mygreen}{2}  & \textcolor{mygreen}{2}  & \textcolor{mygreen}{2}  & \textcolor{mygreen}{2}  & \textcolor{mygreen}{2}  & \textcolor{myblue}{1}   & \textcolor{mygreen}{2}  & \textcolor{mygreen}{2}  & \textcolor{mygreen}{2}  & \textcolor{mygreen}{2}  & \textcolor{myblue}{1}   & \textcolor{myorange}{3} & \textcolor{mygreen}{2}  & \textcolor{myorange}{3}\\
Ottoman    & 0                       & 0                       & \textcolor{myorange}{3} & \textcolor{myorange}{3} & \textcolor{myorange}{3} & 0                       & \textcolor{myorange}{3} & \textcolor{myorange}{3} & 0                       & 0                       & 0                       & 0                       & 0                       & 0                       & 0                       & 0                       & 0                       & 0                       & 0                       & 0\\
Slavonic   & 0                       & 0                       & \textcolor{mygreen}{2}  & 0                       & \textcolor{myblue}{1}   & 0                       & \textcolor{myblue}{1}   & 0                       & 0                       & 0                       & 0                       & 0                       & \textcolor{myorange}{3} & 0                       & \textcolor{myorange}{3} & \textcolor{myblue}{1}   & 0                       & 0                       & \textcolor{myorange}{3} & 0\\
\hline
\end{tabular}
\end{table}

\subsubsection{Intra-rater variability}

\begin{figure} 
	\centering
	\includegraphics[width=0.9\linewidth]{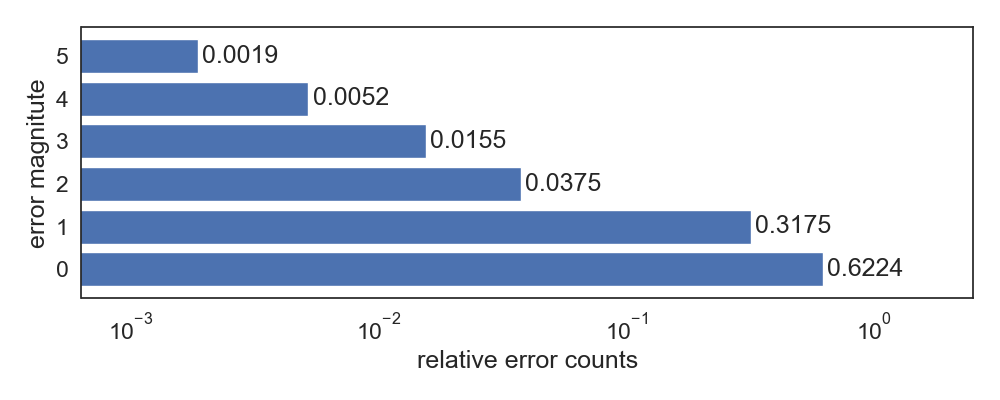}
	\caption{Distribution of absolute intra-rater errors of all participants. The bar width shows the relative frequencies of error magnitudes on a logarithmic scale.}
	\label{fig:intra-obs} 
\end{figure}

To assess intra-rater variability and thus the repeatability of the experiment, the absolute errors between units of duplicate presentations (see Section~\ref{sec:orderofpresentation}) are recorded. 
In accordance with the evaluation strategy given at the beginning of this section, only units with a score greater than zero in at least one of the duplicate presentations are considered. The distributions of absolute errors is visualized in Figure~\ref{fig:intra-obs} on a logarithmic scale. The mean absolute error across all duplicate observations is \textbf{0.469}.

\subsubsection{General inter-rater agreement}

The agreement of different raters (participants) on legibility scores was measured using Intraclass Correlation Coefficients (ICC). Following the definitions of Shrout and Fleiss~\cite{Shrout1979}, we report the ICC variants ICC(2,1), ICC(3,1), ICC(2,k) and ICC(3,k). While the variants ICC(2,x) view the set of participants as a random sample from a larger population and thus express the reliability of the proposed experimental design, variants ICC(3,x) express the reliability of the specific dataset that is published, rated by the specific participants of this study. On the other hand, ICC(x,1) estimate the reliability of a single participant, while ICC(x,k) estimate the reliability of the average of k (in our case k=20) participants~\cite{Shrout1979}. The results are shown in Table~\ref{tab:icc} along with their 95\% confidence intervals.

\begin{table}[!t]
	\renewcommand{\arraystretch}{1.1}
	\caption{Intraclass Correlation Coefficients for all participants}
	\label{tab:icc}
	\centering
	\begin{tabular}{|ll|ll|}
		\hline
		variant & description           & ICC   & 95\% CI       \\
		\hline
		ICC(2,1)    & single random raters  & 0.668 & [0.64, 0.69] \\
		ICC(3,1)    & single fixed raters   & 0.711 & [0.71, 0.71] \\
		ICC(2,k)   & average random raters & 0.976 & [0.97, 0.98] \\
		ICC(3,k)   & average fixed raters  & 0.980 & [0.98, 0.98] \\
		\hline
	\end{tabular}
\end{table}

\subsubsection{Impact of the test environment}

As we work with a mixture of 7 laboratory sessions and 13 online sessions, it is worth assessing the influence of the test environment on rater agreement; uncontrolled factors in the online environment (such as monitor characteristics or insufficient understanding of the task) could lead to a greater divergence between participants.
We thus report ICC(2,1) for the each of the groups separately; other ICC variants are omitted, as we want to address this question independent of a specific set/number of participants. As shown at the top of  Table~\ref{tab:icc-groupwise}, the ICC estimate is indeed higher for the lab participants; however, note the large confidence intervals, where the ICC estimate for online participants is within the 95\% confidence interval of the lab participants.

\begin{table}[!t]
	\renewcommand{\arraystretch}{1.3}
	\caption{Comparison of ICCs of participants grouped by test environment (top) and SMI exposure (bottom)}
	\label{tab:icc-groupwise}
	\centering
	\begin{tabular}{|l|ll|}
		\cline{2-3}
		\multicolumn{1}{c|}{} & ICC(2,1)  & 95\% CI \\
		\hline
		lab participants & 0.702 & [0.65, 0.74] \\
		online participants & 0.664   & [0.63, 0.69] \\
		\hline
		SMI exposure $>$ once per week &  0.690 & [0.66, 0.72] \\
		SMI exposure $\leq$ once per week &  0.649 & [0.57, 0.71]\\
		\hline
	\end{tabular}
\end{table}

\subsubsection{Impact of expertise}
The first hypothesis is that an increased professional exposure to scientific manuscript imagery improves inter-rater agreement. Again, ICC(2,1) are computed separately for participants who work with such imagery multiple times a week and all participants with less exposure (we refrained from a finer distinction due to a small number of participants in the lower exposure categories). The bottom of Table~\ref{tab:icc-groupwise} shows the results. As with the test environment, an effect on the ICC can be observed, but with largely overlapping confidence intervals.

\begin{figure} 
	\centering
	\includegraphics[width=1.0\linewidth]{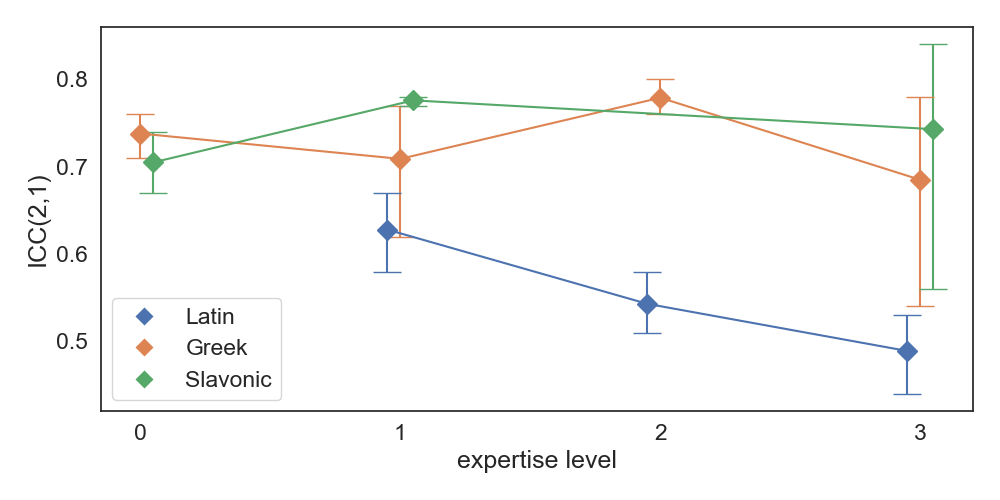}
	\caption{Inter-rater agreement for different levels of expertise in Latin, Greek and Slavonic. Vertical bars show 95\% confidence intervals. Missing values are due to an insufficient number of participants in the respective expertise category (see Table~\ref{tab:participants}.)}
	\label{fig:icc-langskill} 
\end{figure}

Secondly, we investigate if language expertise promotes agreement. For this,  the observations on images containing Latin, Greek and Slavonic (which constitute the bulk of the test images) are treated separately. For each of those image sets, we partition the observations according to the expertise level of their raters in the respective language and compute ICC(2,1) scores. The resulting estimates along with their 95\% confidence intervals are visualized in Figure~\ref{fig:icc-langskill}, where no general trend is observable.

\subsection{Systematic effects and sources of variation}

After addressing the effects of participant and experimental parameters on inter-rater agreement, we now turn to the analysis of their systematic effects, i.e. their influences on mean scores. Furthermore we compare these effects to uncontrolled variations between participants and, most importantly, to the variation introduced by the properties of the observed units.

\begin{figure} 
	\centering
	\subfloat[]{%
		\includegraphics[width=0.24\linewidth]{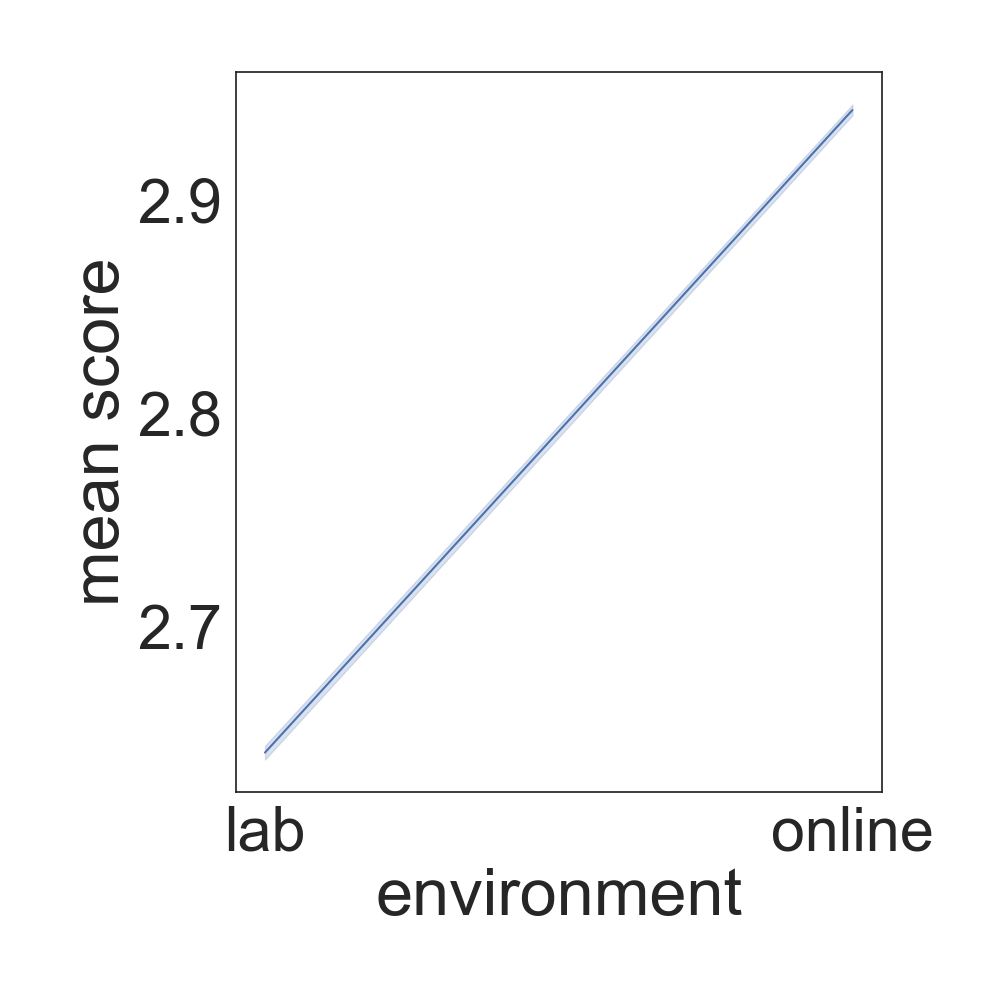}
		\label{fig:meaneffects:line-env}
	}
	\subfloat[]{%
		\includegraphics[width=0.24\linewidth]{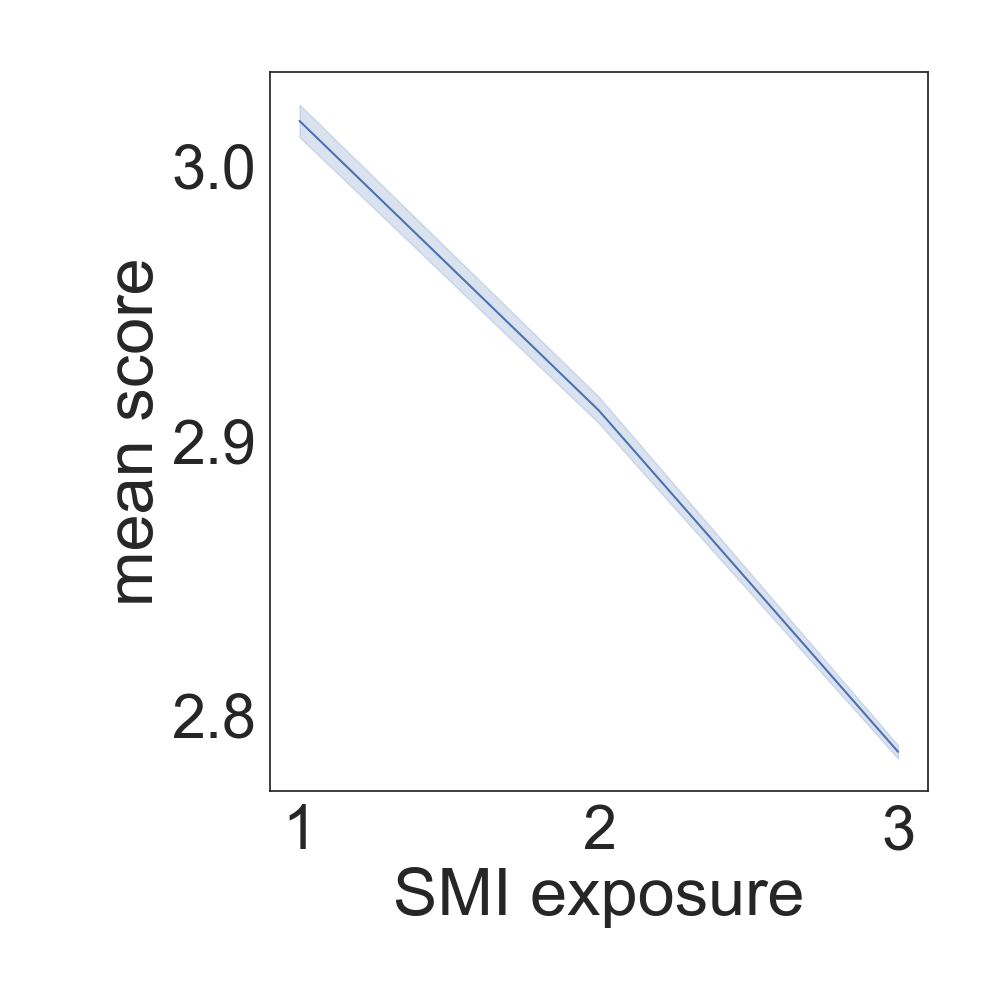}
		\label{fig:meaneffects:line-msiexp}
	}
\subfloat[]{%
	\includegraphics[width=0.24\linewidth]{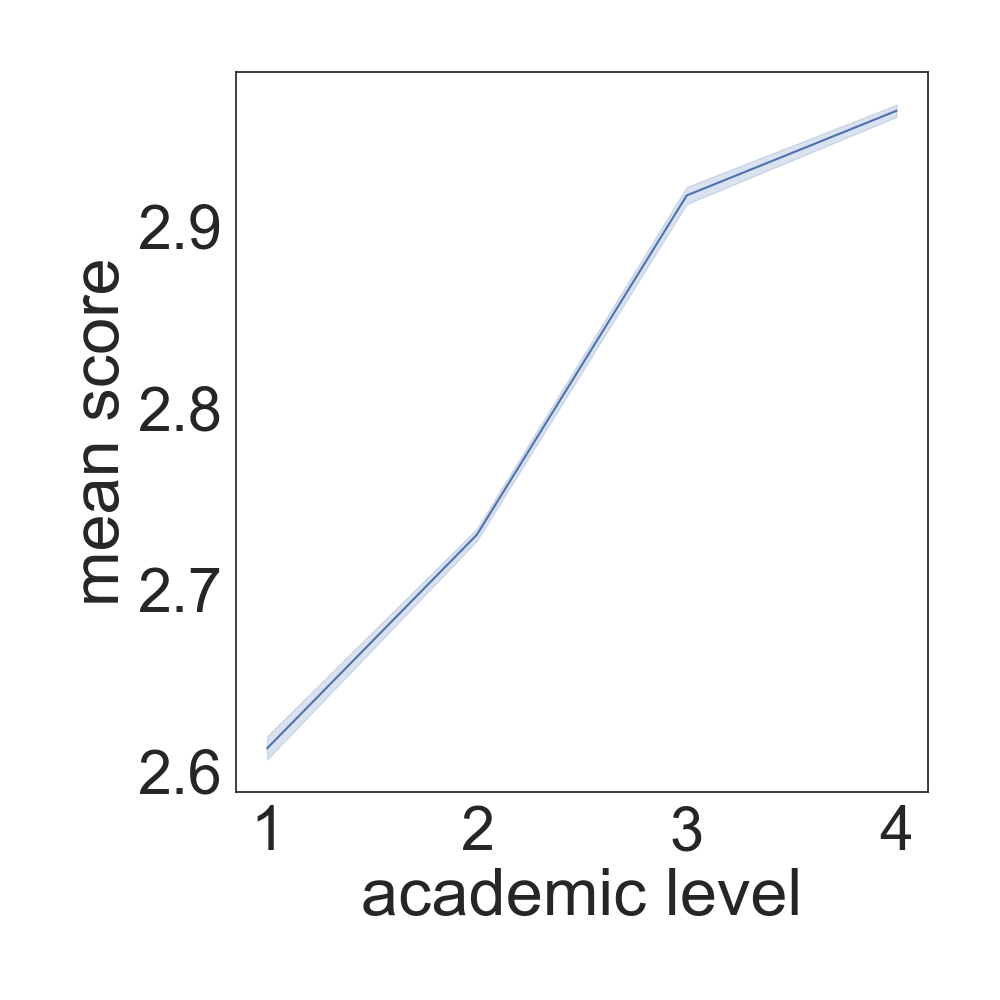}
	\label{fig:meaneffects:line-sclevel}
}
\subfloat[]{%
	\includegraphics[width=0.24\linewidth]{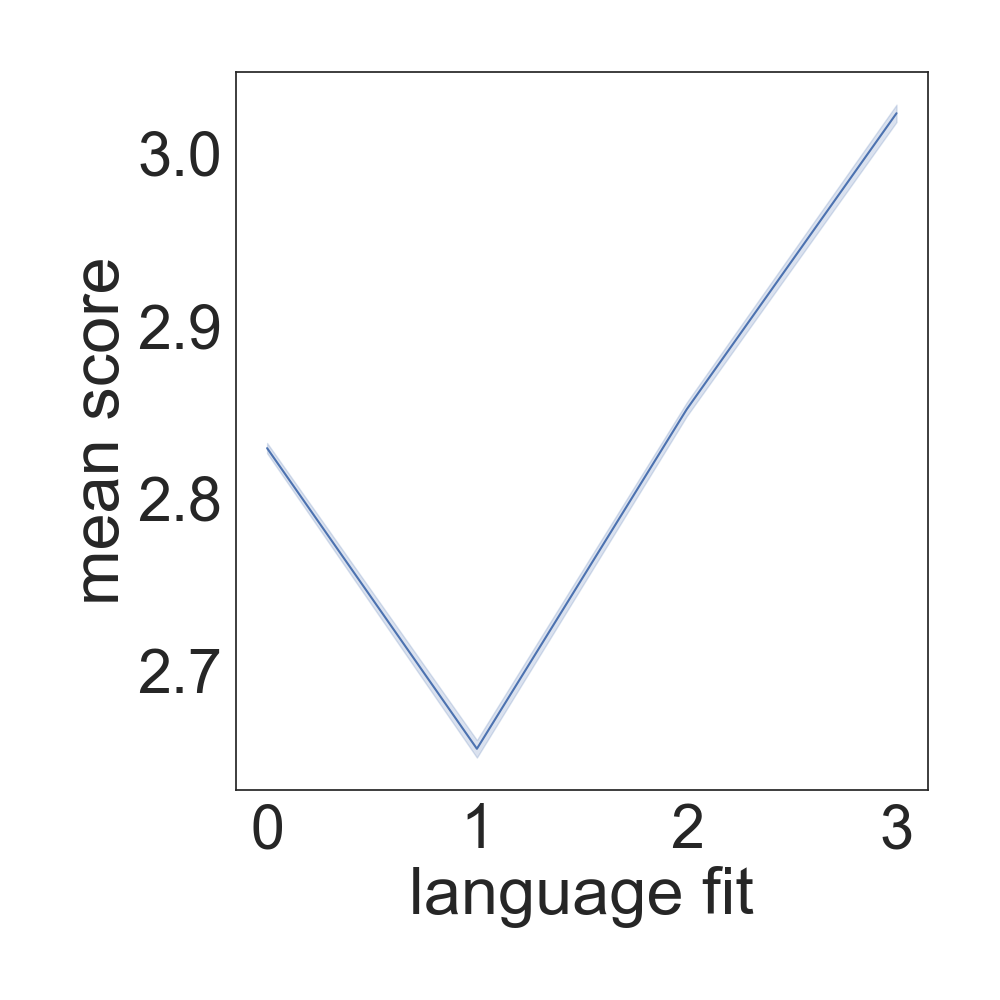}
	\label{fig:meaneffects:line-langfit}
}\\
\subfloat[]{%
	\includegraphics[width=1.0\linewidth]{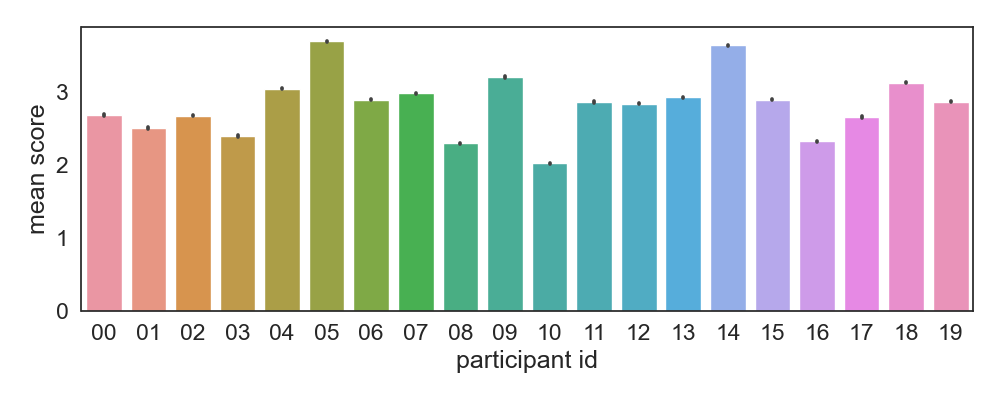}
	\label{fig:meaneffects:mean-per-participant}
}
	\caption{Effects of participant parameters on mean scores: (a) test environment, (b) exposure to scientific manuscript images, (c) academic level, (d) language fit, i.e. for a given observation, the expertise of the participant in the language of the observed manuscript patch. In (e) the mean score for each participant is shown.}
	\label{fig:meaneffects}
\end{figure}

We concentrate on the following controlled parameters which gave reason to suspect linear relations (see Figures~\ref{fig:meaneffects:line-env}-\ref{fig:meaneffects:line-langfit}): test environment, SMI exposure, academic level and language fit. The language fit of an observation is defined as the participant's skill level in the language associated with the observed unit. For the possible values of the above-mentioned parameters refer to Table~\ref{tab:ratingoptions}. The sum of those influences plus an uncontrolled (random) source of variation constitutes the participant-wise variation of means shown in Figure~\ref{fig:meaneffects:mean-per-participant}.

In order to jointly model and investigate those participant-specific effects and the effects of observed units, we employ linear mixed effects models. In our initial model, legibility score is the dependent variable. Fixed effects are test environment, SMI experience, academic level and language fit. As further uncontrolled variations between participants are to be expected, we include the participant ID as a random intercept. A second random intercept is defined as the ID of the observed unit to model the dependence of an observation's score on the observed unit. The model was fitted in \textit{R}~\cite{R} using the \textit{lme4} package~\cite{lme4}.

For each of the fixed effects a likelihood ratio test of the full model against a model without the respective effect was performed. We found that language fit affects legibility scores by an increase of \textbf{0.05} per skill level, at a $p<0.001$ confidence level. For the other fixed effects, p values are above 0.05.

The random effects 'unit ID' and 'participant ID' contribute a variance of \textbf{2.133} and \textbf{0.136}, respectively; the residual variance is \textbf{0.803}.

\subsection{Spatial distribution of variability}

Finally, we investigate if the spatial distribution of units with a high variability in legibility scores follows a pattern. Qualitative inspection of standard deviation maps that are part of the published dataset (see Figure~\ref{fig:dataset:std} for an example) suggests that the highest variability is found near the boundaries of text areas and is thus caused by the variations in bounding box placement. 

\begin{figure} 
	\centering
	\subfloat[]{%
		\includegraphics[width=0.35\linewidth]{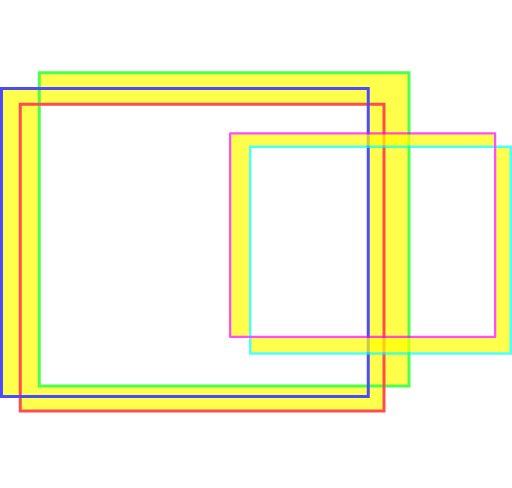}
		\label{fig:borderareas:illu}
	}
	\subfloat[]{%
		\includegraphics[width=0.65\linewidth]{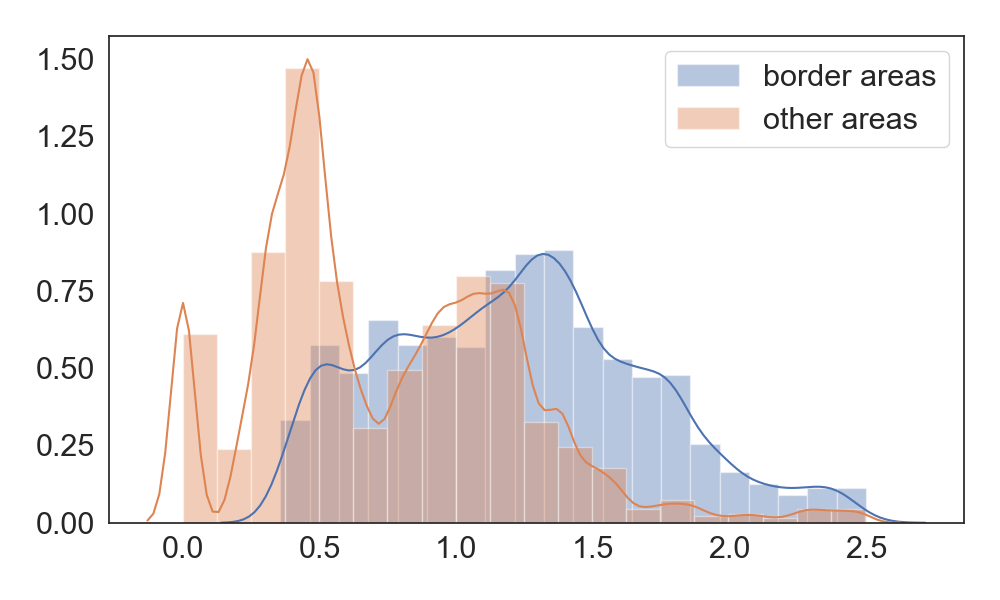}
		\label{fig:borderareas:std-comp}
	}
	\caption{(a) Definition of critical border areas. The differently colored rectangles are bounding boxes defined by different participants. The union of pairwise symmetric differences of similar bounding boxes (in terms of intersection over union) is shown in yellow. (b) Distribution of standard deviations for border areas versus other areas.}
	\label{fig:borderareas}
\end{figure}

To test this hypothesis, we define such critical border areas as the union of symmetric differences between pairs of similar bounding boxes from different participants; in this context, we consider bounding boxes as similar if their intersection-over-union ratio greater than 0.9. The idea is illustrated in Figure~\ref{fig:borderareas:illu}.

It was found that 7.56\% of units fall under the definition of border areas given above. The distributions of standard deviations of border areas and other areas are shown in Figure~\ref{fig:borderareas:std-comp}. The mean standard deviation of border areas (\textbf{1.218}) is significantly higher than the mean standard deviation of other areas (\textbf{0.750}) at a $p<0.001$ confidence level (according to a Mann-Whitney-U test as the respective distributions are non-normal).

\section{Dataset description and validity}
\label{sec:dataset}

\begin{figure} 
	\centering
	\subfloat[]{%
		\includegraphics[width=0.33\linewidth]{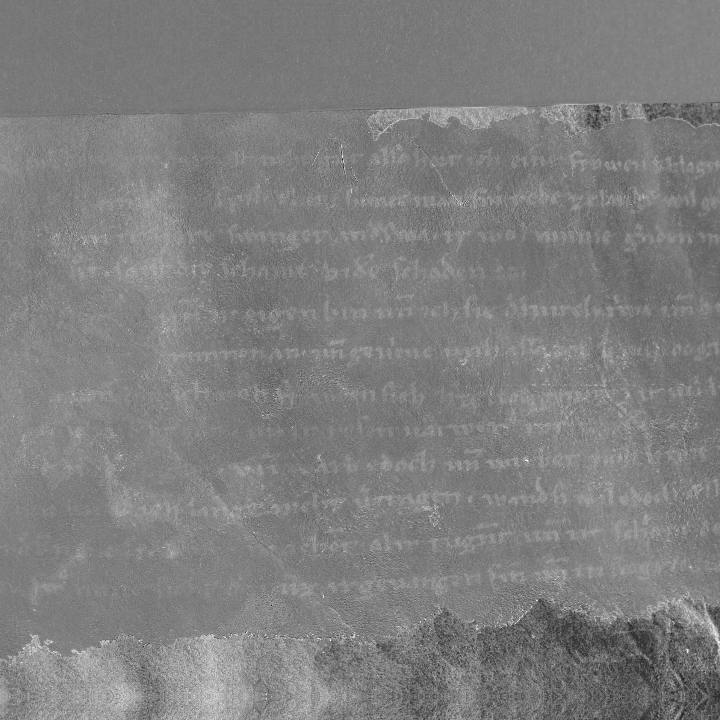}
		\label{fig:dataset:img}
	}
	\subfloat[]{%
		\includegraphics[width=0.33\linewidth]{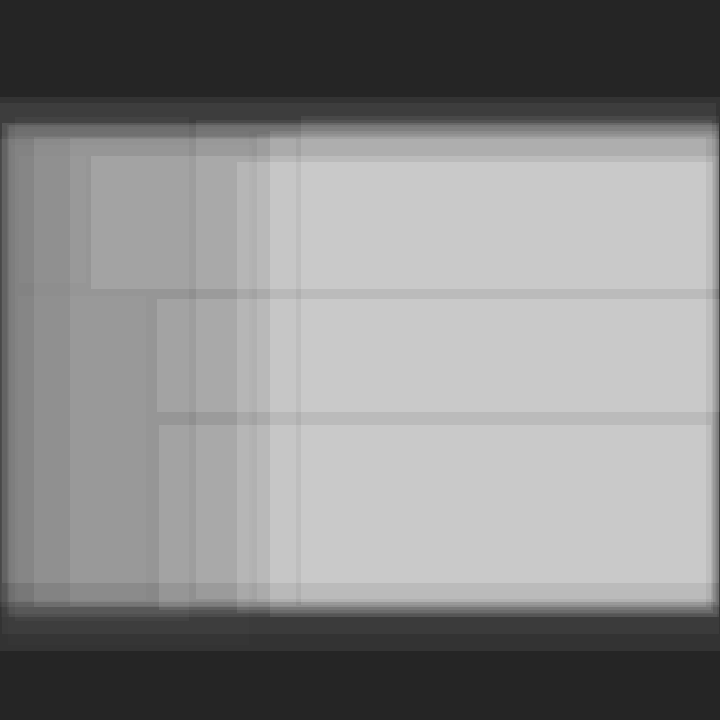}
		\label{fig:dataset:score}
	}
	\subfloat[]{%
		\includegraphics[width=0.33\linewidth]{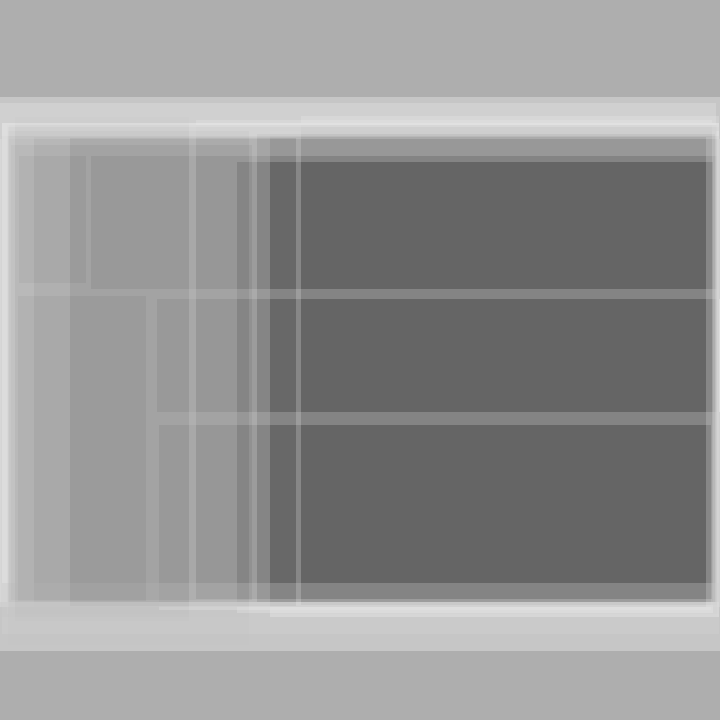}
		\label{fig:dataset:std}
	}
	\caption{An instance of the dataset: (a) test image; (b) mean score map; (c) standard deviation map.}
	\label{fig:dataset} 
\end{figure}

As described in Section~\ref{sec:testimages}, the SALAMI dataset is based on 250 test images: 50 manuscript regions are represented by 5 processed variants each. Along with every test image, we publish a legibility map averaging the scores of all participants as well as a standard deviation map, showing the spatial distribution of uncertainty. See Figure~\ref{fig:dataset} for an example of such an image triplet. The test images together with their legibility maps are ready to be used as a ground truth for developing computer vision methods for the localized estimation of legibility in manuscript images. Additionally to the estimation of absolute legibility from a single image, the dataset also supports pairwise comparison or ranking applications due to the 5 variants per manuscript region contained in the dataset.
The standard deviation maps can be used to exclude regions with a high uncertainty from training/evaluation, or for implementing weighted loss function depending on local uncertainty.

Additionally to the pixel maps, the originally recorded data (.json encoded) are provided, along with well-documented python scripts that can be used to reproduce the legibility and score maps as well as the results described in Section~\ref{sec:evaluation}. The dataset, code and documentations are available on zenodo~\cite{salami_zenodo}.

As the SALAMI dataset is the first of its kind, we have to justify its validity as well as the appropriateness of the novel study design used to generate it. For this purpose we summarize the results of the statistical analyses obtained in Section~\ref{sec:evaluation}.
\paragraph{Repeatability}
Even with background areas excluded, 62.2\% of absolute intra-rater errors are zero and the mean absolute error is 0.469.
\paragraph{Reliability}
Table~\ref{tab:icc} summarizes the inter-rater agreement and thus reliability of the test method (ICC(2,x)) as well as the specific test results published in the form of our dataset (ICC(3,x)). According to the highly cited guidelines by Koo et al.~\cite{Koo2016}, we can in either case expect moderate reliability from a single rater (ICC(x,1)) but excellent reliability from the average of 20 independent raters (ICC(x,k)). We have not found conclusive evidence for a negative impact of online participation or lack of participant skills/experience on the reliability.
\paragraph{Systematic effects}
The linear mixed effects model analysis does not show a statistically significant impact of SMI experience, test environment or academic level on the scores obtained. Only the property of language fit has a significant effect; however, we would not attribute any practical relevance to an increase of mean scores by 0.05 for each language skill level. Furthermore it should be noted that the language skills of participants were acquired through self-assessment and not validated in any way.
\paragraph{Sources of variation}
The second and most important conclusion that can be drawn from the linear mixed effects model is that the identity of the observed unit contributes the majority of variance in the scores (2.133), while the identity of the participant only contributes a variance of 0.136. This is a good indicator that the measured legibility is an objective property of a given image area and not dominated by subjective preferences of the participants.
\paragraph{Spatial distribution of uncertainty}
We show that uncertainty is varying spatially within test images and that especially border regions of text areas exhibit an increased variability in scores. This motivates the publishing of mean score maps along with standard deviation maps as an essential part of the dataset.

\section{Conclusion}
\label{sec:conclusion}

We have conducted a study with 20 experts of philology and paleography to create the first dataset of Subjective Assessments of Legibility in Ancient Manuscript Images, intended to serve as a ground truth for developing and validating computer vision-based methods for quantitative legibility assessment. Such methods, in turn, would elevate a whole research field centered around the digital restoration of written heritage; their development is the subject of future work.

Additionally to creating the dataset itself, we describe a novel methodology to conduct similar studies in the future, demonstrating the validity of the results and the robustness against variations in test environment and participant properties.


We have collected qualitative comments of the participating experts regarding their perspective on the study design, which revealed potentials for improvement in future work. Specifically, the following issues were pointed out:
\begin{itemize}
	\item Line height has an impact on readability. This property of the assessed images is not considered in our analysis.
	\item The study only assesses the percentage of text that is readable; however, another relevant dimension would be the time/effort required to read a text.
	\item Manuscript experts tend to dynamically 'play' with the images (i.e. vary contrast, brightness, scale, etc.) in order to decipher texts. This was not permitted in our test design, potentially biasing the results.
\end{itemize}

\section*{Acknowledgment}

The authors would like to thank all study participants who enabled this research by contributing their valuable time and expertise.
This research was funded by the Austrian Science Fund (FWF) under grant No P29892.

{\small
	\bibliographystyle{IEEEtran}
	\bibliography{references}
}


\end{document}